\documentclass[10pt,letterpaper]{article}
\usepackage[top=0.85in,left=2.75in,footskip=0.75in]{geometry}

\usepackage{amsmath,amssymb}

\usepackage{changepage}

\usepackage[utf8x]{inputenc}

\usepackage{textcomp,marvosym}

\usepackage{cite}

\usepackage{nameref,hyperref}

\usepackage{microtype}
\DisableLigatures[f]{encoding = *, family = * }

\usepackage[table]{xcolor}

\usepackage{array}

\usepackage{float}

\newcolumntype{+}{!{\vrule width 2pt}}

\newlength\savedwidth

\raggedright
\usepackage[aboveskip=1pt,labelfont=bf,labelsep=period,justification=raggedright,singlelinecheck=off]{caption}

\makeatletter
\renewcommand{\@biblabel}[1]{\quad#1.}
\makeatother

\date{}

\usepackage{lastpage,fancyhdr,graphicx}
\usepackage{epstopdf}
\fancyheadoffset[L]{2.25in}
\fancyfootoffset[L]{2.25in}

\begin{document}
\vspace*{0.2in}

\begin{flushleft}
{\Large
\textbf\newline{Modeling Retinal Ganglion Cell Population Activity with Restricted Boltzmann Machines} 
}
\newline
\\
Matteo Zanotto\textsuperscript{1},
Riccardo Volpi\textsuperscript{1},
Alessandro Maccione\textsuperscript{2},
Luca Berdondini\textsuperscript{2},
Diego Sona\textsuperscript{1},
Vittorio Murino\textsuperscript{1},
\\
\bigskip
\textbf{1} Pattern Analysis and Computer Vision (PAVIS), Istituto Italiano di Tecnologia, Genova, Italy
\\
\textbf{2} Neuroscience and Brain Technology (NBT), Istituto Italiano di Tecnologia, Genova, Italy
\\
\bigskip

%

\end{flushleft}
\section*{Abstract}
The retina is a complex nervous system which encodes visual stimuli before higher order processing occurs in the visual cortex. In this study we evaluated whether information about the stimuli received by the retina can be retrieved from the firing rate distribution of Retinal Ganglion Cells (RGCs), exploiting High-Density 64x64 MEA technology. To this end, we modeled the RGC population activity using mean-covariance Restricted Boltzmann Machines, latent variable models capable of learning the joint distribution of a set of continuous observed random variables and a set of binary unobserved random units. The idea was to figure out if binary latent states encode the regularities associated to different visual stimuli, as modes in the joint distribution. We measured the goodness of mcRBM encoding by calculating the Mutual Information between the latent states and the stimuli shown to the retina. Results show that binary states can encode the regularities associated to different stimuli, using both gratings and natural scenes as stimuli. We also discovered that hidden variables encode interesting properties of retinal activity, interpreted as population receptive fields. We further investigated the ability of the model to learn different modes in population activity by comparing results associated to a retina in normal conditions and after pharmacologically blocking GABA receptors ($GABA_C$ at first, and then also $GABA_A$ and $GABA_B$). As expected, Mutual Information tends to decrease if we pharmacologically block receptors. We finally stress that the computational method described in this work could potentially be applied to any kind of neural data obtained through MEA technology, though different techniques should be applied to interpret the results.\\~\\ 
\textbf{\textit{This is an early version of the paper, a further revision is ongoing and this version will be updated when finalized.}}



\section*{Introduction}
Retinal Ganglion Cells play a key role in the visual system, being the only connection between the retina and the areas of the brain dedicated to the process of the visual  information. For this reason, understanding how the visual information is stored in the RGC neural activity is a very interesting neuroscientific problem. Multi-Electrode Array (MEA) Technology \cite{mea12009} represents a novel and unique opportunity to study neural tissues, but defining novel techniques to make sense of the high dimensional output (spike trains) is an active problem. In this work we show that information related to the visual stimuli shown to the retina can be recovered from the RGC activity by modeling the population distribution with a probabilistic graphical model, through a novel approach that scales very well both in terms of the number of neurons recorded and in time. We also address the problem of extracting information related to population activities, with the same approach, showing that our model can learn latent representations of neural signals, such as population receptive fields.\\ Different works addressed the problem of decoding RGC activity. Bialek et al. \cite{dec01991} first introduced the problem of decoding the activity of a movement-sensitive neuron of the fly visual system. Warland et al. \cite{dec1997} showed that stimuli (in their case, photopic, spatially uniform, temporally broadband flicker) can be decoded from spike trains, and understood that decoding performance highly improved considering many spike trains. Pillow et al. \cite{dec2008} evaluated a decoding approach based on correlations among different neurons. Marre et al. \cite{dec2015} defined a linear decoder trained over a population of RGCs stimulated with a moving bar, and were able to decode the position of the bar from neural activity with high accuracy. Botella-Soler et al. \cite{dec2016} addressed the problem of reconstructing the stimulus proposed to RGCs from spike activities, reporting promising results. A different area of research is related to infer the latent structure of RGCs. For example, Linderman et al. \cite{linderman2016} showed that type of cells and receptive field center locations can be unsupervisely learnt from spike trains, using a Generalized Linear Model \cite{nelder1972}. Our goal was different from the ones proposed in literature, since we were trying to unsupervisely extract information about the stimuli proposed, not building a decoder to reconstruct them. When we discuss about \textit{latent representations}, it must not be confused with latent strutures discussed in \cite{linderman2016}, since the latent representations introduced in this work represent modes in the firing rate distribution.\\ To carry out the experiments, we acquired different datasets, obtained through High Density $64\times64$ MEA Technology, coupled with a projector showing stimuli to the retina on the chip. For our analysis we are thus considering activities of hundreds of neurons. In this work, we are making the assumption that the visual information is encoded in the spiking rate of the RGCs, following the rate coding approach. This choice was led by the fact we believe modeling the spike rate is more informative than modelling spike co-occurrence. We inferred the firing rates from the spike trains using log-Gaussian Cox processes \cite{lgcp1998}. This choice was led by the fact that these models deal with uncertainty at different levels of abstraction, and so they well apt to the stochasticity of the neural activity. At this point, we needed a model able to find the regularities that - we hypothesize - are related to the visual stimuli. To this end, we chose to model the joint distribution of the firing rates with mean-covariance Restricted Boltzmann Mahchines, latent variable models capable of learning the joint distribution of a set of continuous observed random variables and a set of binary unobserved random variables (often referred to as latent variables). Once the model is trained, this bi-partition allows to represent regularities in the observed inputs through the inferred value of the latent binary variables. As a measure of the goodness of the modeling, we calculated the mutual information between the binary states and the different stimuli. First, we performed a two-hour experiment where square gratings with 800 µm-wide bars moving in 8 directions (0° to 360°) were used. After having validated the assumption that the regularities encoded by the different states are associated to different stimuli, we ran a test to evaluate the physiological plausibility of our findings. The experiment consisted in recording the RGC activity of an adult mouse retina (P45) when stimulated with three different types of moving square gratings (0.011, 0.023 and 0.045 cycle/degree), both in normal conditions and blocking $GABA_A$ and $GABA_B$ receptors. As expected, Mutual Information decreased after having blocked the receptors, because pharmacologically induced impairment led to to an unreliable population code. Finally, we evaluated whether natural stimuli can be retrieved from the firing rate distribution, using neural data from another experiment. The paper is organized as follows: in the following section the Methods are described, namely the different protocols used and the mathematical tools; the Results are reported in the next section; the last section includes the Conclusion, where also future work is detailed.

\section*{Materials and Methods}

%

\subsection*{Data Acquisition}

In this section, the different stimulation protocol and experimental modalities defined to validate our approach are detailed. All experiments were carried out with High Density $64 \times 64$ MEA Technology. Channels were arrenged on a $64 \times 64$ lattice and separated by $42 um$. Sampling rates were in the range of $7-8 kHz$, stimuli were shown to the retina with a frequency of $30 Hz$. The detailed description of the system is reported in \cite{mea12009,mea22014}. Figure 1 shows an overview of the experimental setup.

\begin{figure}
\centering
\includegraphics{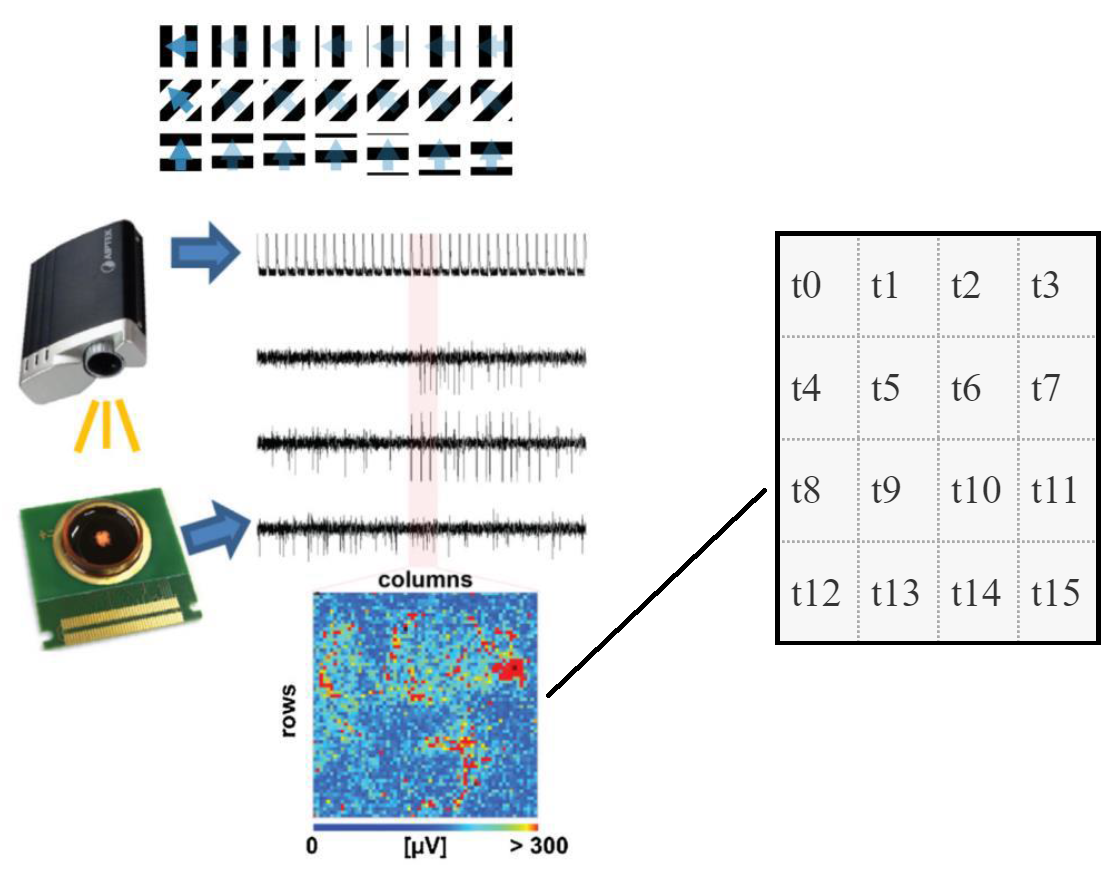}
\caption{Setup of the system used to record spikes from RGCs. Different stimuli were shown to the retina through a projector. After the experimental sessions, we were able to associate each firing rate sample to the stimulus displayed at that time. We refer division of the chip in patches (t0 - t15) in the Result section.}\label{fig:}
\end{figure}

\subsubsection*{Heterogeneous Grating Stimulation}

We stimulated a P38 stage mouse's retina with an heterogeneous stimulation protocol. The idea was to evaluate if specific features shared by different images are encoded in specific retinal activation modes. Specifically, a moving square grating  with 800 µm-wide bars was presented to the retina in 8 different orientations: 0°, 45°, 90°, 135°, 180°, 225°, 270° and 315°. For each orientation, ten repetitions of the stimuli were carried out. Figure 2 (\textit{left}) shows different samples of stimuli. 
\begin{figure}
\centering
\includegraphics[width=\textwidth]{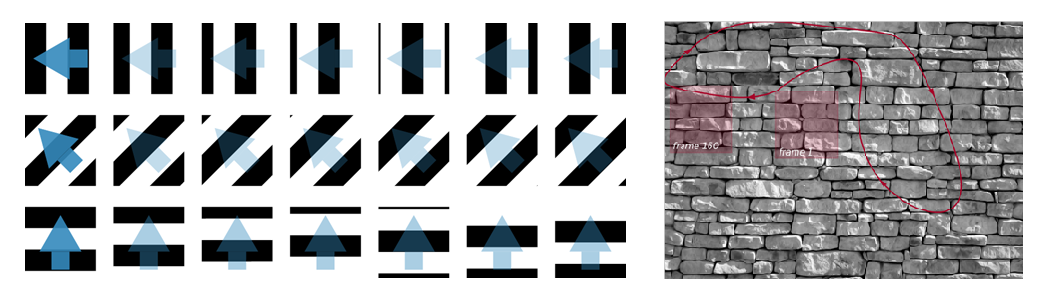}
\caption{Left: heterogeneous grating stimulation. Right: trajectory followed to show natural scene stimuli to the retina.}\label{fig:}
\end{figure}

\subsubsection*{GABA Blockage}

We recorded the activity of an adult mouse retina (P45 stage) when stimulated with three different moving square gratings  in three different recording conditions. At first, the retina in normal conditions was subjected to the three following stimuli:
\begin{enumerate}
\item Square grating at 0.011 cycles/degree
\item Square grating at 0.023 cycles/degree
\item	 Square grating at 0.045 cycles/degree
\end{enumerate}
All the stimulation protocols had temporal frequency of $1 Hz$, Michelson contrast 0.5 and mean luminance $1.36 cd/m^2$. The following stages of the experiment consisted in repeating the same stimulations under two different pharmacological treatments aimed at progressively blocking $GABA$ receptors and hence impairing the functional behaviour of the inhibitory circuitry. Specifically, $TPMPA$ $150 µM$ was added first to block $GABA_C$ receptors and Bicuculline $10 µM$ was successively added to also block $GABA_A$ and $GABA_B$ receptors. The rationale behind the experiment was recording the activity of a retina having a progressively more impaired inhibitory circuitry and checking whether the decreased spatio-temporal precision of the encoding expected from physiology [ref?] was captured by the mc-RBM.

\subsubsection*{Natural scenes}

Our last eperiment was defined to understand whether we can retrieve the regularities associated to more complex stimuli, such as natural scenes. To this end, a pilot study was performed by projecting on the retina a sequence of frames generated by scanning natural images following a smooth closed trajectory. Figure 2 (\textit{right}) shows an example where the photo of a brick wall was used to extract the stimulation sequence. The idea behind this stimulation protocol was trying to simulate a smooth scanning movement over a natural scene and see how the retina reacted to it.


\subsection*{Mean-Covariance Restricted Boltzmann Machines}
\begin{figure}
\centering
\includegraphics[width=0.6\textwidth]{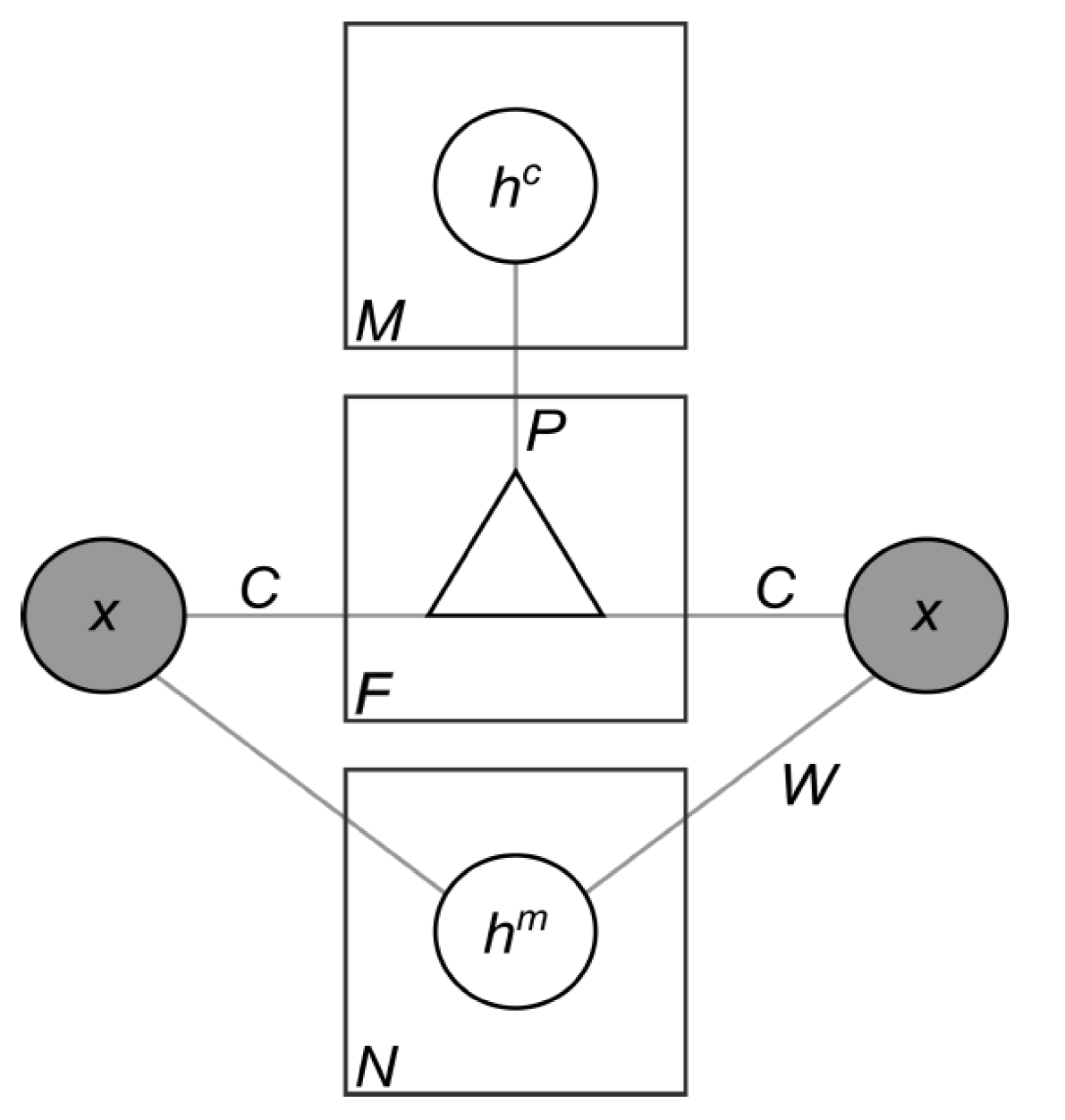}
\caption{Mean-covariance Restricted Boltzmann Machine \cite{mcrbm_ranzato2010,mcrbm_dahl2010}. The details of the model are reported in text.}\label{fig:}
\end{figure}
Restricted Boltzmann Machines (RBMs) \cite{rbm1986} are latent variable models (two-layer networks) capable of learning the joint distribution of a set of observed random variables and a set of binary unobserved random variables (often referred to as latent variables), in an unsupervised fashion. In this model, the input variables are connected with stochastic, binary feature detectors, through symmetric weights. The technical report by Hinton \cite{hinton2010} gives a detailed explaination of this model, in this section we report only the details necessary to a basic understanding of it. RBMs are energy models, and the energy associated to the joint configuration of visible and hidden units, in case of binary input and hidden units, is defined by:
\begin{equation}
E(\textbf{v},\textbf{h}) = - \textbf{a}^T\textbf{v} - \textbf{b}^T\textbf{h} - \textbf{v}^T\textbf{W}\textbf{h} 
\end{equation}
where \textbf{v},\textbf{h} are the values associated to visible and hidden units, respectively; \textbf{a},\textbf{b} are the biases associated to those units and \textbf{W} is the weight matrix. The energy is related to the probability density of visible and hidden units via
\begin{equation}
P(\textbf{v},\textbf{h}) \propto e^{-E(\textbf{v},\textbf{h})}
\end{equation}
In case we want to model real-valued inputs, visible units are replaced with linear units with independent Gaussian noise, and the energy function becomes
\begin{equation}
E(\textbf{v},\textbf{h}) = - \frac{1}{2}(\textbf{v} - \textbf{a})^T(\textbf{v} - \textbf{a}) - \textbf{c}^T\textbf{h} - \textbf{v}^T\textbf{W}\textbf{h}
\end{equation}
assuming unitary standard deviation of the Gaussian noise associated to visible units. Mean-covariance RBMs \cite{mcrbm_ranzato2010,mcrbm_dahl2010} are a modified version of the standard model where the hidden units are divided in two different sets: mean units and precision units. The energy function is thus divided in two different terms: $E_{mc}= E_m + E_c$, where the former is defined by (3) and the latter is defined by
\begin{equation}
E_c(\textbf{v},\textbf{h}_c) = -\textbf{d}^T\textbf{h}_c - (\textbf{v}^T\textbf{R})^2\textbf{P}\textbf{h}_c
\end{equation}
where \textbf{R} is the visible-factor weight matrix, \textbf{P} the factor hidden pooling matrix, and \textbf{d} the hidden bias vector. The conditional distribution of the hidden units given the visibles is given by
\begin{equation}
P(\textbf{h}\arrowvert\textbf{v}) = \sigma\Big(\textbf{d} + \big((\textbf{v}^T\textbf{R}^2\textbf{P})\big)^T\Big)
\end{equation}
The conditional distribution of the visible units given the hiddens is given by
\begin{equation}
P(\textbf{v}\arrowvert\textbf{h}_c,\textbf{h}_m) \propto N(\Sigma\textbf{W}\textbf{h}_m,\Sigma)
\end{equation}
where
\begin{equation}
\Sigma = \big(\textbf{R}(diag\big(-\textbf{P}^T\textbf{h}_c)\big)\textbf{R}^T\big)^{-1}
\end{equation}
We can unsupervisely train such energy-based models using a fast algorithm, \textit{contrastive divergence} \cite{cd2000}. For a detailed explaination of the learning procedure we recommend to read \cite{hinton2010}. Once the model is trained, this bi-partition allows to represent regularities in the observed inputs through the inferred value of the latent binary variables. An important fact is that conditioning on the latent variables, the observed ones are approximately jointly Gaussian distributed \cite{mcrbm_ranzato2010}. The mean and covariance of these Gaussians are defined by the specific values of the formers. This means that each of the possible binary vectors representing the latent variable values is associated to one mode of the joint distribution of the inputs. This aspect makes mc-RBMs a very good model for our purposes of finding the regularities associated to visual stimuli shown to the retina. One of these sets is used to model the mean value of the inputs, the other is thought for explicitly modelling the covariance between the observed variables. The presence of this second set allows to obtain a much better fit of the data distribution than what can be achieved with simpler models like Gaussian Restricted Boltzmann Machines.\\
In our framework, each single unit corresponds to a different neuron, and the values fed in input are firing rate values at different time steps. To train the model, we need to give in input a matrix where the rows are the $N_s$ values of firing rates and the columns are associated to the $N_n$ neurons recorded. Following the usual way of training RBMs, we shuffled the rows of our dataset and divided it in mini-batches of size $100$: during the training procedure, weights were updated after the model has passed through one mini-batch, averaging over the gradients calculated for each sample of the mini-batch.
 
\subsection*{Model evaluation}
It can be difficult to interpret what RBMs have learnt, because we lack of a direct measure that defines how the model is fitting our data.
For this work, we relied on information theory techniques to evaluate how well models learnt, and defined a visualization technique inspired by Spike Trigger Average \cite{sta2001} to understand \textit{what} the models had learnt. Specifically, to understand whether the modes encoded by mc-RBMs are effectively associated to particular visual stimuli, we decided to exploit Mutual Information (MI). It provides a measure about how mutually dependent two continuous or discrete random variables are: in our case, we calculated it between the stimuli poposed to the retina and the binary states learnt by the mc-RBMs. MI for discrete random variables is defined by the following equation
\begin{equation}
I(X;Y) = \sum_{y \in Y}\sum_{x \in X}p(x,y)log\Big(\frac{p(x,y)}{p(x)p(y)}\Big)
\end{equation}
where, in our case, $x$ and $y$ are associated to different stimuli and different binary states. We could easily perform this step, because we knew from the experiment protocol the stimuli associated to every firing rate sample. MI provides a measure (a real value in the range $0-1$) which tells us how well the model has encoded the regularities associated with the visual stimuli. MI can give us a general idea about the learning outcome and can be really helpful to compare different models. However, it does not carry any information about \textit{what} the model has learnt. To this end, we defined a technique to understand the meaning of the different binary states, in terms of visual stimuli encoded. Such method is closely related to STA, and was implemented as follows: (i) one binary state is selected; (ii) the firing rate samples associated to that particular state are retrieved; (iii) the stimuli associated to those samples are averaged. Following this procedure, we can understand whether the regularities encoded by the mc-RBM states are associated to different stimuli. This technique can also be applied to the single binary units of the model, selecting one binary variable and then following steps (ii) and (iii).

\section*{Results}
Results show that information related to visual stimuli shown to the retina can be retrieved from the joint distribution of RGC firing rates. We proved it using mc-RBMs applied to firing rates inferred through log-Gaussian Cox processes. In addition we show that, while the binary states of the machines are associated to regularities related to visual stimuli, the single binary units encode interesting firing rate activities, that we interpreted as population receptive fields. To train the models, we used a variation of the contrastive divergence algorithm, the \textit{persistent contrastive divergence} \cite{pcd2008}. In order to model only neurons that carry information relative to the stimuli, we chose to use for our experiments only neurons with a time-varying spike-rate. To this end, we calculated the maximum and the minimum values of the derivative of the firing rate for each neuron, removing the cells when the difference between the maximum and the minimum was below a threshold (arbitrarialy set to $10^{-8}$). In this way, we are removing not only silent neurons, but also neurons with a static, banal firing rate, namely a firing rate that surely does not carry any information about the stimuli. To distinguish the different cells, we performed Spike Sorting before inferring the firing rates, using Plexon Offline Sorter \cite{plexon} with the option t-dist EM. The following section is organized as follows: in the first section the results obtained through the visualization techniques described in previous section are reported, using data acquired during the grating experiment; in the following section, the results associated to the GABA blockage experiment are detailed; in the last section, the results obtained from the experiment with natural scenes are detailed.

\subsection*{Decoding visual stimuli from RGC firing rates}

\begin{figure}
\centering
\includegraphics[width=12cm, height=12cm]{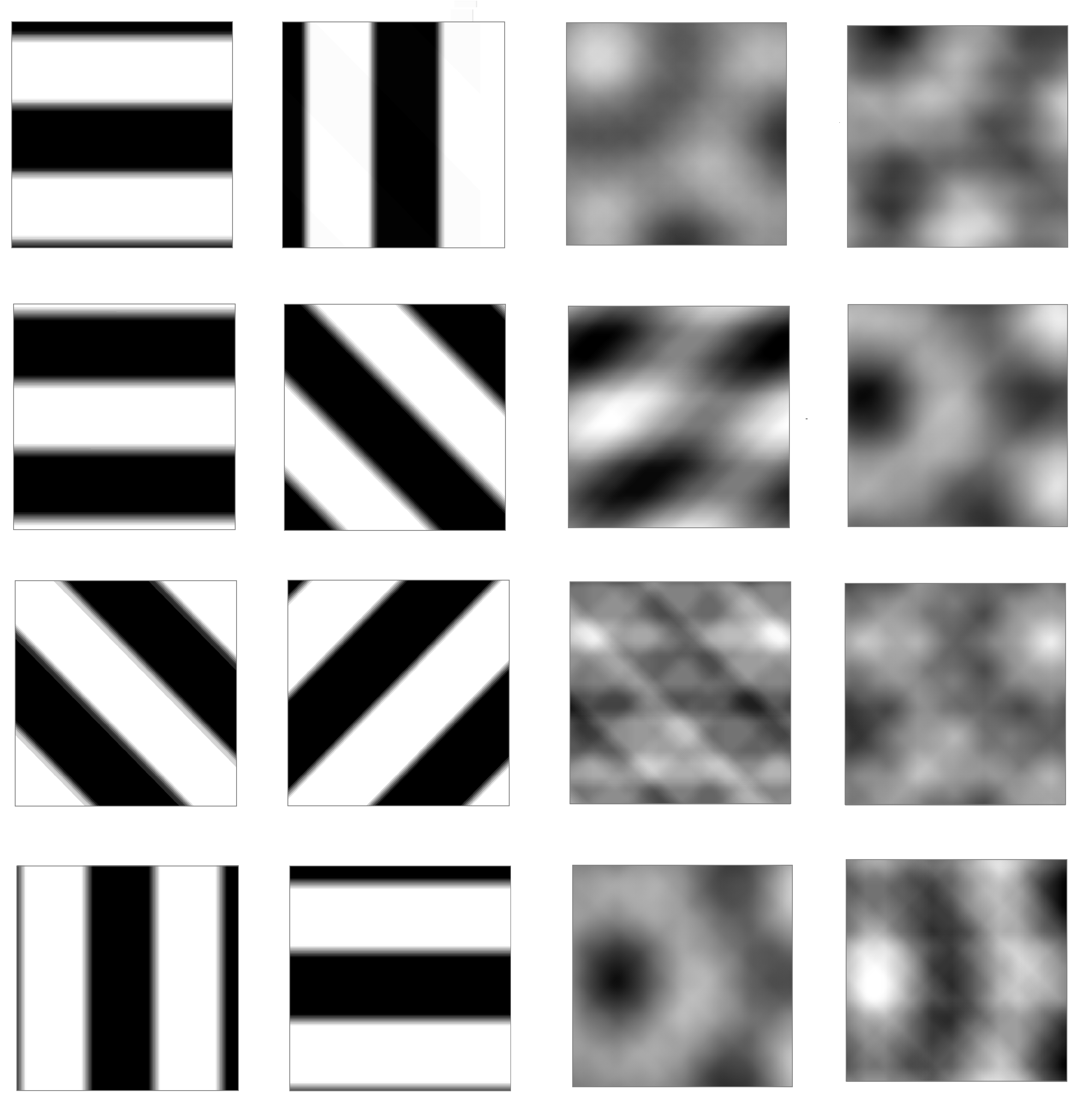}
\caption{Each row is associated with a different patch of the MEA chip, from which neurons were recorded during the experiments. Top-down: patches t0, t4, t7, t8. The first two columns represent averages over stimuli associated with different \textit{binary states}, the last two columns represents averages over stimuli associated with different \textit{binary units}. }\label{fig:}
\end{figure}
The results reported in this section were achieved using data acquired with the modalities previously described, with a retina (stage P38) stimulated with heterogeneous gratings. In order to manage the complexity of the task, the chip was divided in 16 non-overlapping regions ($16\times16$ electrodes each) and the recorded data were processed by 16 different mcRBMs (shown in Figure 1). This was nececessary due to the memory limitations of the GPU used for the analysis, but is also in line with what happens in the retina where local signal correlations play an important role in encoding the visual input. Totally, 3923 neurons were recorded, but only 1344 were used for the experiments, following the selection described. Figure 4 shows the results obtained: the images reported were obtained following the technique described. Each row is associated to neurons recorded from different patches of the chip. From top to bottom: t0 (114 neurons), t4 (132 neurons), t7 (61 neurons), t8 (104 neurons). The first two columns contain images associated with different \textit{binary states}, the last two columns contain images associated with different \textit{binary units}. The interpretation of the results reported in the first two columns is straightforward: latent states encode regularities in RGC joint distribution associated to different phases of the stimuli proposed to the retina. This is a key result in our study, since the main goal was to figure out if such information could be recovered from the population activity. As can be observed, averaged pictures are smoother than original stimuli: this is a positive aspect of the results, since we can expect the population activity associated to one stimulus to be more similar to the activities associated to comparable stimuli, rather than activities associated to very different stimuli. The consequence of this similarity is that firing rate samples associated to similar stimuli are encoded in the same binary states. We observed that increasing the size of the mc-RBM hidden state, averages were less smooth. This is an obvious consequence, since the machine can learn up to $2^N$ different states, where $N$ is the number of hidden units. If $N$ is big, the model can learn a number of regularities big enough that it does not need to aggregate different activities in the same state. The interpretation of the results reported the last two columns is more difficult. Our explanation is that hidden units encode population receptive fields. With this consideration, the interpretation of the results would be that while hidden \textit{states} encode regularities associated to \textit{particular stimuli} shown to the retina, hidden \textit{variables} encode regularities associated to biological properties of the retina, such as population receptive fields. It would be interesting to evaluate if these receptive fields are approximation of the ones of Lateral Geniculate Nucleus cells, since each LGN cell receives inputs from different RGCs, and in this framework we are considering ensemble of cells.   

\subsection*{Effects of GABA blockage on the model}

After having validated that states learnt by the mc-RBM are associated to modes in the RGC joint distribution closely related to the stimuli proposed, we were interested in evaluating how induced impairment in the inhibitory circuitry is reflected in the latent states. As expected, GABA blockage results in a lower MI between the states and the stimuli. Results are reported in Figure 5, which pictures the MI scores associated to the three cases described, with three different types of grating stimuli. Figure 6 shows the distribution of the stimuli associated to an arbitrarily chosen state. The distribution related to the state learnt \textit{before} pharmacological treatment is associated to a precise range of stimuli, as expected from results shown in previous section; the distribution of stimuli associated to a state learnt \textit{after} the treatment is highly corrupted, due to the fact that retinal behaviour is abnormal, which results in an unreliable population code. 
\begin{figure}
\centering
\includegraphics[width=8cm, height=6cm]{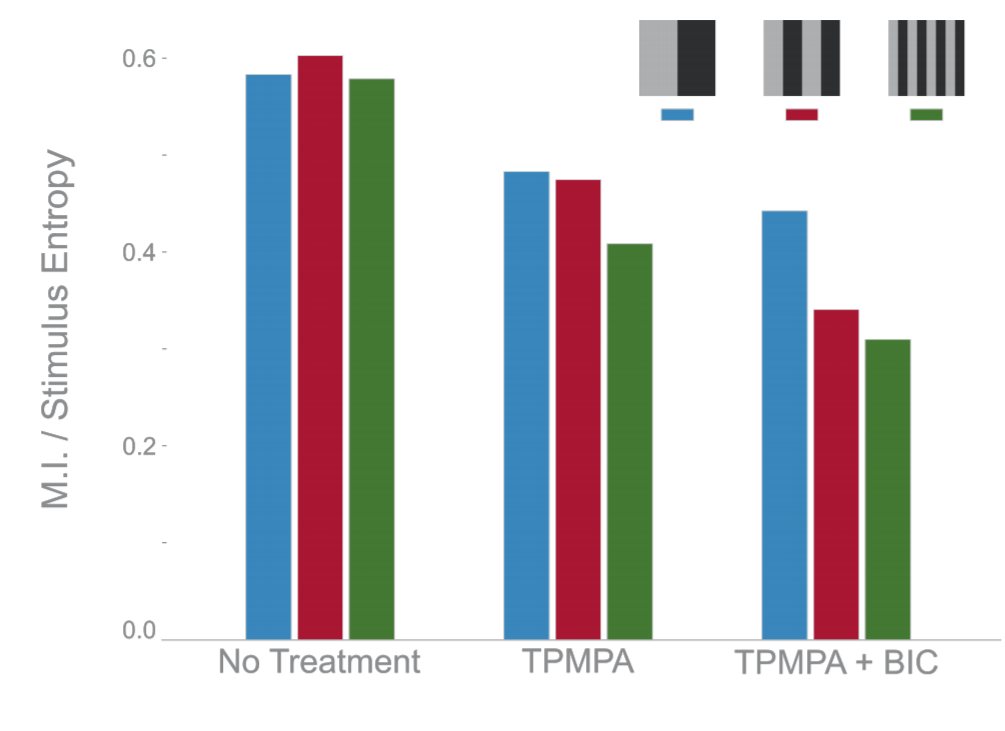}
\caption{Mutual Information associated to (left to right) retina without GABA blockage, retina with $GABA_c$ blockage and retina with also $GABA_a$ and $GABA_b$ blockage. Different colors are associated to gratings with different spatial frequency.}\label{fig:}
\end{figure}
\begin{figure}
\centering
\includegraphics[width=12cm, height=6cm]{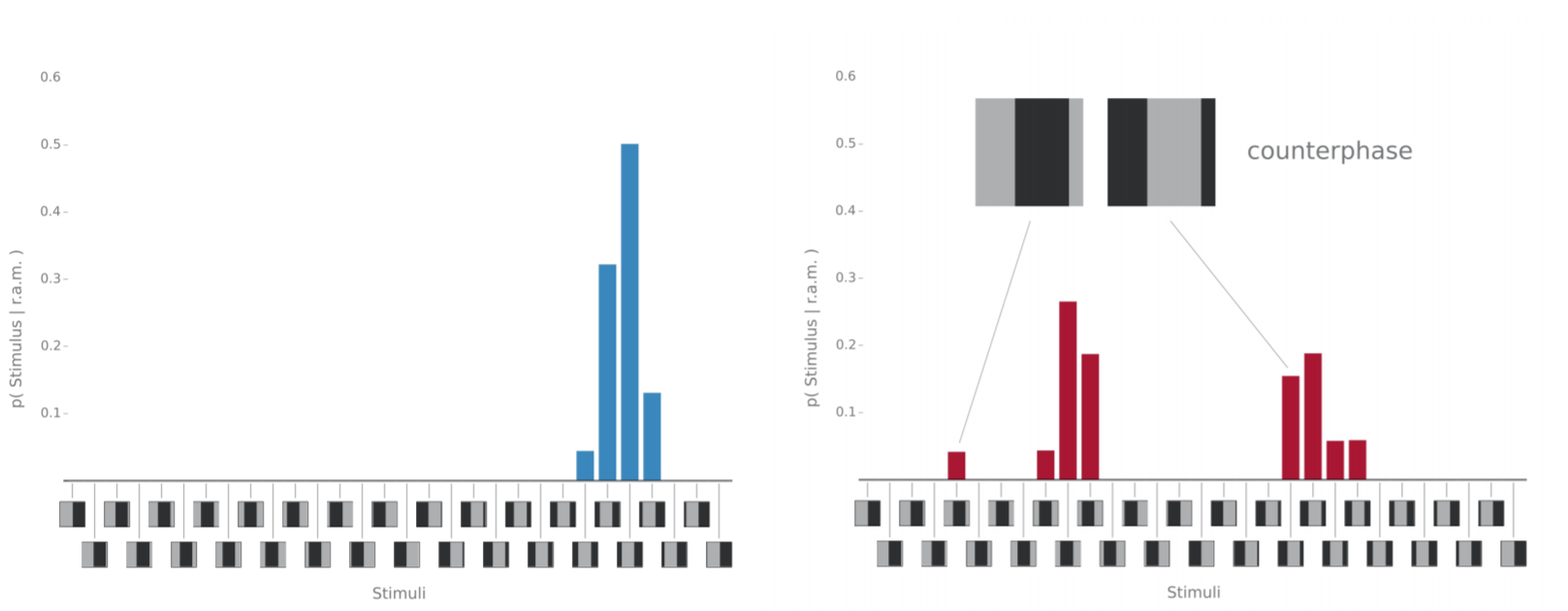}
\caption{Distribution of the stimuli associated to a single latent state. While in retina without impairment a good match is found (left), blocking GABA receptors results in a mismatch (right).}\label{fig:}
\end{figure}

\subsection*{Natural scenes}

Finally, we were interested to understand if RGCs produce a reliable response also when stimulated with natural scenes (specifically, a video showing frames of a wall). In case of a reliable response, we believed that our model could capture the modes in the distribution also in this scenario. Results proposed in this section were achieved using a P? stage mouse's retina, and the model was fed with the activities of neurons from t5. The findings reported in Figure 7 are consistent with the findings of the gratings experiment: latent states encode the regularities associated with the stimuli proposed. With this kind of stimuli, latent variables encode smoother versions of the stimuli.    
\begin{figure}
\centering
\includegraphics[width=1.0\textwidth]{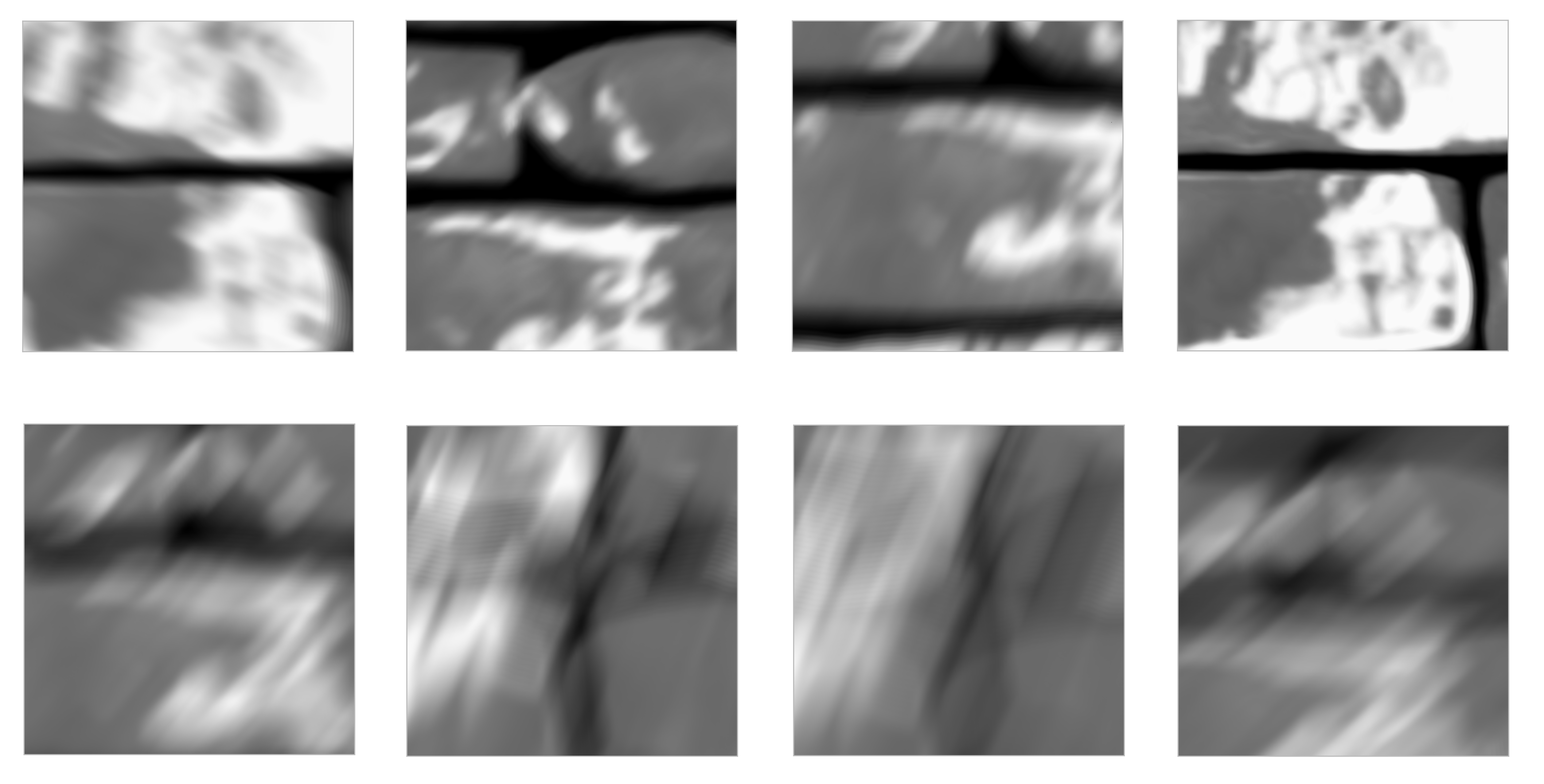}
\caption{Top: stimuli encoded by different binary states learnt by the model. Bottom: visual features encoded by different binary units. The model was trained over firing rate recorded during the experiment with natural scenes.}\label{fig:}
\end{figure}

%
%


\section*{Conclusion}

We proposed a new method to decode population activity and extract latent representations of the neuronal signals. Such method can potentially scale to thousands of neurons and to very long temporal sequences, using state-of-the-art GPUs. We evaluated this approach with different protocols: the first experiment, with heterogeneous gratings showed us that stimuli can be retrieved from RGC activities and interesting properties sush as population receptive fields can be detected; the pharmacological blockage of GABA receptors was a validation step; finally, the natural scene (wall) experiment was carried out to understand if more complex stimuli could be retrieved from the signals. Further investigations will be made in order to better understand the properties learnt by the binary variables, through a fine analysis of the type of cells active in the different activations. We are also interested in evaluating if we can retrieve temporal evolution of different stimuli from the joint distribution of RGCs. Some preliminary results are reported in \cite{nips2016}.

\section*{Acknowledgments}
This research received financial support from the 7th Framework Programme for Research of the European Commission, under Grant agreement no. 600847: RENVISION project of the Future and Emerging Technologies (FET) programme Neuro-bio-inspired systems (NBIS) FET-Proactive Initiative.


%
%
%


\begin{thebibliography}{10}

\bibitem{gollisch2013}
Gollisch T.
\newblock Nonlinear Spatial Integration in the Receptive Field Surround of Retinal Ganglion Cells.
\newblock Journal of Phisiology-Paris, Volume 107, Issue 5, November 2013, Pages 338–348.

\bibitem{takeshita2014}
Takeshita D and Gollisch T.
\newblock Features and Functions of Nonlinear Spatial Integration by Retinal Ganglion Cells.
\newblock The Journal of Neuroscience, 28 May 2014, 34(22): 7548-7561.

\bibitem{cd2000}
Hinton GE.
\newblock Training products of experts by minimizing contrastive divergence.
\newblock Technical Report GCNU TR 2000-004, Gatsby Unit, University College London, 2000.

\bibitem{crbm2006}
Taylor G et al.
\newblock Modeling Human Motion using Binary Latent Variables.
\newblock Advances in Neural Information Processing Systems (NIPS), 2006.

\bibitem{mcrbm_ranzato2010}
Ranzato M and Hinton GE. 
\newblock Modeling Pixel Means and Covariances Using Factorized Third-Order Boltzmann Machines. 
\newblock In Computer Vision and Pattern Recognition, 2010

\bibitem{mcrbm_dahl2010}
Dahl GE et al.
\newblock Phone recognition with the mean-covariance restricted Boltzmann machine.
\newblock Advances in Neural Information Processing Systems (NIPS), 2010.

\bibitem{lgcp1998}
Møller J. et al.
\newblock Log Gaussian Cox processes.
\newblock Scandinavian Journal of Statistics, 25(3):451–482, 1998.

\bibitem{sta2001}
Chichilnisky EJ.
\newblock A simple white noise analysis of neuronal light responses.
\newblock Network: Computation in Neural Systems, 12:199-213, 2001.

\bibitem{mea12009}
Berdondini L et al.
\newblock Active pixel sensor array for high spatio-temporal resolution electrophysiological recordings from single cell to large scale neuronal networks“
\newblock Lab on a Chip, 9, 2644 - 2651, (2009)

\bibitem{mea22014}
Maccione A et al.
\newblock Following the Ontogeny of Retinal Waves: Pan-Retinal Recordings of Population Dynamics in the Neonatal Mouse
\newblock The Journal of Physiology, 592, 1545-1563 (2014)

\bibitem{dec01991}
Bialek W et al.
\newblock Reading a neural code.
\newblock Science. 1991 Jun 28;252(5014):1854-7.

\bibitem{dec1997}
Warland DK et al. 
\newblock Decoding visual information from a population of retinal ganglion cells. 
\newblock J Neurophysiol. Nov;78(5):2336–50 (1997).

\bibitem{dec2008}
Pillow JW et al.
\newblock Spatio-temporal correlations and visual signalling in a complete neural population. 
\newblock Nature 454: 995–9 (2008).

\bibitem{dec2015}
Marre O et al. 
\newblock High accuracy decoding of a dynamical motion from a large retinal population. 
\newblock PLOS Comput Biol 11: e1004304 (2015).

\bibitem{dec2016}
Botella-Soler V et al.
\newblock Nonlinear decoding of a complex movie from the mammalian retina.
\newblock arXiv:1605.03373 [q-bio.NC] (2016)

\bibitem{linderman2016}
Linderman S et al.
\newblock Bayesian latent structure discovery from multi-neuron recordings 
\newblock Advances in Neural Information Processing Systems 29 (2016)

\bibitem{rbm1986}
Smolensky P.
\newblock Information processing in dynamical systems: Foundations of harmony theory. 
\newblock In Parallel Distributed Processing: Explorations in the Microstructure of Cognition, vol. 1: Foundations, pp. 194–281. MIT Press (1986)

\bibitem{hinton2010}
Hinton GE. 
\newblock A Practical Guide to Training Restricted Boltzmann Machines.
\newblock Technical Report UTML TR 2010–003, University of Toronto, 2010.

\bibitem{nips2016}
Volpi R et al.
\newblock Unsupervised Learning of Spatio-Temporal Features from Retinal Neuronal Signals.
\newblock NIPS Workshop, Brains and Bits: Neuroscience Meets Machine Learning (2016).

\bibitem{nelder1972}
Nelder JA.
\newblock Generalized linear models. 
\newblock J. R. Stat. Soc. A 135, 370–384 (1972).

\bibitem{pcd2008}
Tieleman T.
\newblock Training restricted Boltzmann machines using approximations to the likelihood gradient. 
\newblock In: Proceedings of the Twenty-first International Conference on Machine Learning (ICML 2008). ACM (2008)

\bibitem{plexon}
http://www.plexon.com/products/offline-sorter.

\end{thebibliography}
\end{document}